\documentclass[runningheads]{llncs}
\usepackage{graphicx}
\usepackage{multirow}
\usepackage{xcolor}
\usepackage{subfig}
\usepackage{comment}

\usepackage{amsmath}
\usepackage{gensymb}
\usepackage{makecell}

\begin{document}
\title{Place recognition in gardens by learning visual representations: data set and benchmark analysis}
\titlerunning{Place recognition in gardens by learning visual representations}
%
\author{Mar\'ia Leyva-Vallina  \and
Nicola Strisciuglio \and
Nicolai Petkov}
\authorrunning{M. Leyva-Vallina, N. Strisciuglio, N. Petkov}
%
\institute{Bernoulli Institute for Mathematics, Computer Science and Artificial Intelligence, \\ University of Groningen, the Netherlands
\\
\email{m.leyva.vallina@rug.nl}}
\maketitle              
\begin{abstract}
Visual place recognition is an important component of systems for camera localization and loop closure detection. It concerns the recognition of a previously visited place based on visual cues only. 
 Although it is a widely studied problem for indoor and urban environments, the recent use of robots for automation of agricultural and gardening tasks has created new problems, due to the challenging appearance of garden-like environments. Garden scenes predominantly contain green colors, as well as repetitive patterns and textures. The lack of available data recorded in gardens and natural environments makes the improvement of visual localization algorithms difficult.

In this paper we propose an extended version of the TB-Places data set, which is designed
 for testing algorithms for visual place recognition. It contains images with ground truth camera pose recorded in real gardens in different seasons, with varying light conditions. We constructed and released a ground truth for all possible pairs of images, indicating whether they depict the same place or not.

We present the results of a benchmark analysis of  methods based on convolutional neural networks for holistic image description and place recognition. We train existing networks (i.e. ResNet, DenseNet and VGG NetVLAD) as backbone of a two-way architecture with a contrastive loss function. The results that we obtained demonstrate that learning garden-tailored representations contribute to an improvement of performance, although the generalization capabilities are limited.
\keywords{Benchmarking  \and data set \and deep learning \and place recognition.}
\end{abstract}
\section{Introduction}

Visual place recognition is a widely studied problem in Computer Vision and concerns the recognition of a previously seen scene based on the analysis of visual features only~\cite{Lopez-Antequera2017,Lowry2016}. The problem gained great interest among researchers in the fields of robotics and computer vision due to its applications to autonomous driving~\cite{McManus2014}, robot navigation~\cite{Cummins2008,Milford2012,Geiger2013,Sunderhauf2013,Lopez-Antequera2017}, camera localization based on image retrieval~\cite{sattler2018benchmarking} and loop-closure detection~\cite{sunderhauf2011brief}.  It is a key component for visual localization based on image retrieval algorithms. Given a query image, a reference image with known camera pose that depicts the same place has to be retrieved from a database. Subsequently, the relative pose between query and reference image can be calculated, and the camera pose in the reference system corresponding to the query image can be estimated. In this work, treat place recognition as distinguishing between pairs of similar and dissimilar images.

Visual place recognition algorithms face challenges when appearance changes occur due to variations of illumination, weather, season, camera viewpoint or when repetitive textures are present~\cite{Torii2015pami}. However, depending on the particular environments where place recognition algorithms are applied, the specific problem and challenges can vary. For instance, in outdoor environments, large changes in illumination (day and night)~\cite{Sattler2012,Milford2012} and weather conditions (sun and rain)~\cite{Sunderhauf2013} are often present. In addition,  urban scenes are subject to partial or total occlusions due to obstacles (i.e. vehicles or pedestrians), while countryside environments are affected by seasonal changes~\cite{Badino2011,Sunderhauf2013}. 

Holistic image descriptors were proposed to face these challenges, such as SeqSLAM~\cite{Milford2012,Sunderhauf2013} and NetVLAD~\cite{Arandjelovic2017}. The former performs place recognition by matching sequences of images, while the latter is a CNN-based method that employs a new orderless pooling VLAD layer, trained with weakly-labeled image triplets (one reference, a set of positive matches, and a set of negative matches). Outdoor environments are usually large, therefore high spatial precision for recognition or localization is not required. In contrast, indoor environments are typically smaller, and a more accurate localization is necessary. In addition, indoor place recognition methods are required to be able to recognize a scene even under substantial viewpoint variations, and are usually faced with repetitive patterns~\cite{Shotton2013}. Algorithms based on local features like 
FAB-MAP have been successfully implemented for visual localization in indoor environments~\cite{Cummins2008}~\cite{Cummins2009}.

With the raise of interest in gardening and agricultural robotics~\cite{bac2014harvesting,ohi2018design,walter2018flourish,strisciuglio2018trimbot2020}, new challenges and problems have become relevant for place recognition algorithms. Gardens are affected by illumination and seasonal changes, and localization algorithms are required to also be robust to viewpoint variations, while faced with very similar and repetitive textures. Moreover, gardens have a lot of internal similarity, i.e. a bush can look similar to all the other bushes in the garden. Thus, for a localization algorithm to be successful, it is required to capture and describe the relevant parts of the scene and their relative arrangement, while ignoring irrelevant elements (i.e. the common background). 

In~\cite{Leyva2019}, we released the first version of the TB-Places data set for bechmarking the performance of existing algorithms in garden environments, and recorded low recognition results. Existing algorithms and CNN-based models for place recognition are not robust to the challenges provided by garden-like environments. Thus, more data and garden-specialized place recognition methods  are needed to advance the state-of-the-art of computer vision applied to gardening and agricultural robotics. 

In this paper, we propose an extended version of the TB-Places data set, with more than 23K new images. We learn garden-specific feature descriptors by training several models of deep Convolutional Neural Networks as backbone for a siamese architecture with a contrastive loss function. We carried out experiments with ResNet~\cite{he2016deep}, DenseNet~\cite{huang2017densely} and a VGG~\cite{Simonyan2014} pre-trained with a VLAD~\cite{Arandjelovic2017} layer as backbone for the place recognition architecture. The results that we obtained demonstrate that learning garden-specific representations contributes better recognition, but the generalization capabilities are still limited.

The paper is structured as follows. We introduce the extended version of the TB-Places data set in Section~\ref{sec:dataset}. In Section~\ref{sec:experiments}, we explain the network architecture that we employ to compute the image descriptors and the training procedure, while in Section~\ref{sec:results} we present and compare the results that we achieved. Finally, we draw conclusions in Section~\ref{sec:conclusions}.

\section{Data set}
\label{sec:dataset}

\begin{figure}[!t]
\centering
        
\subfloat[]{
\includegraphics[width=0.35\textwidth]{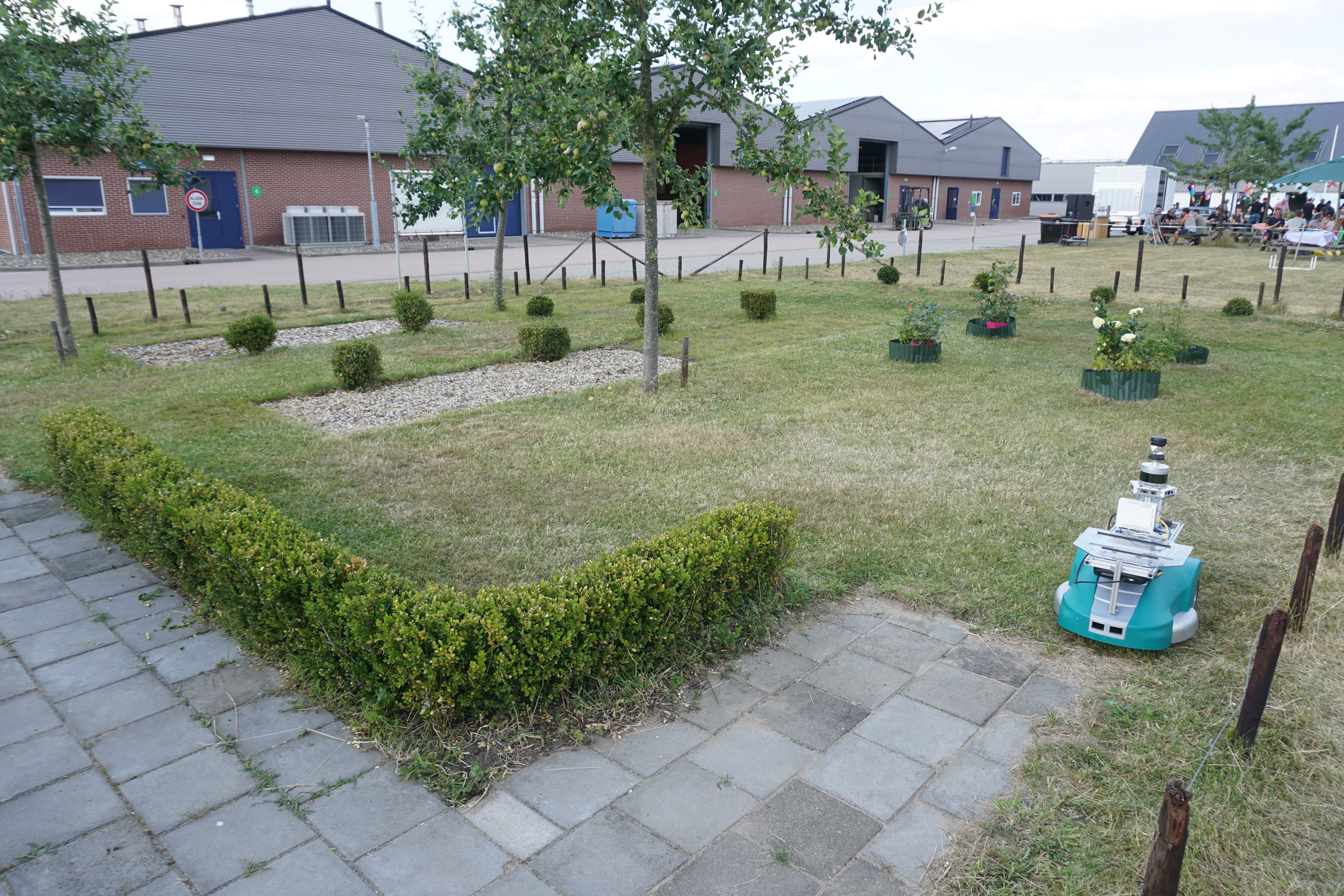}}
~\subfloat[]{\includegraphics[width=0.35\textwidth]{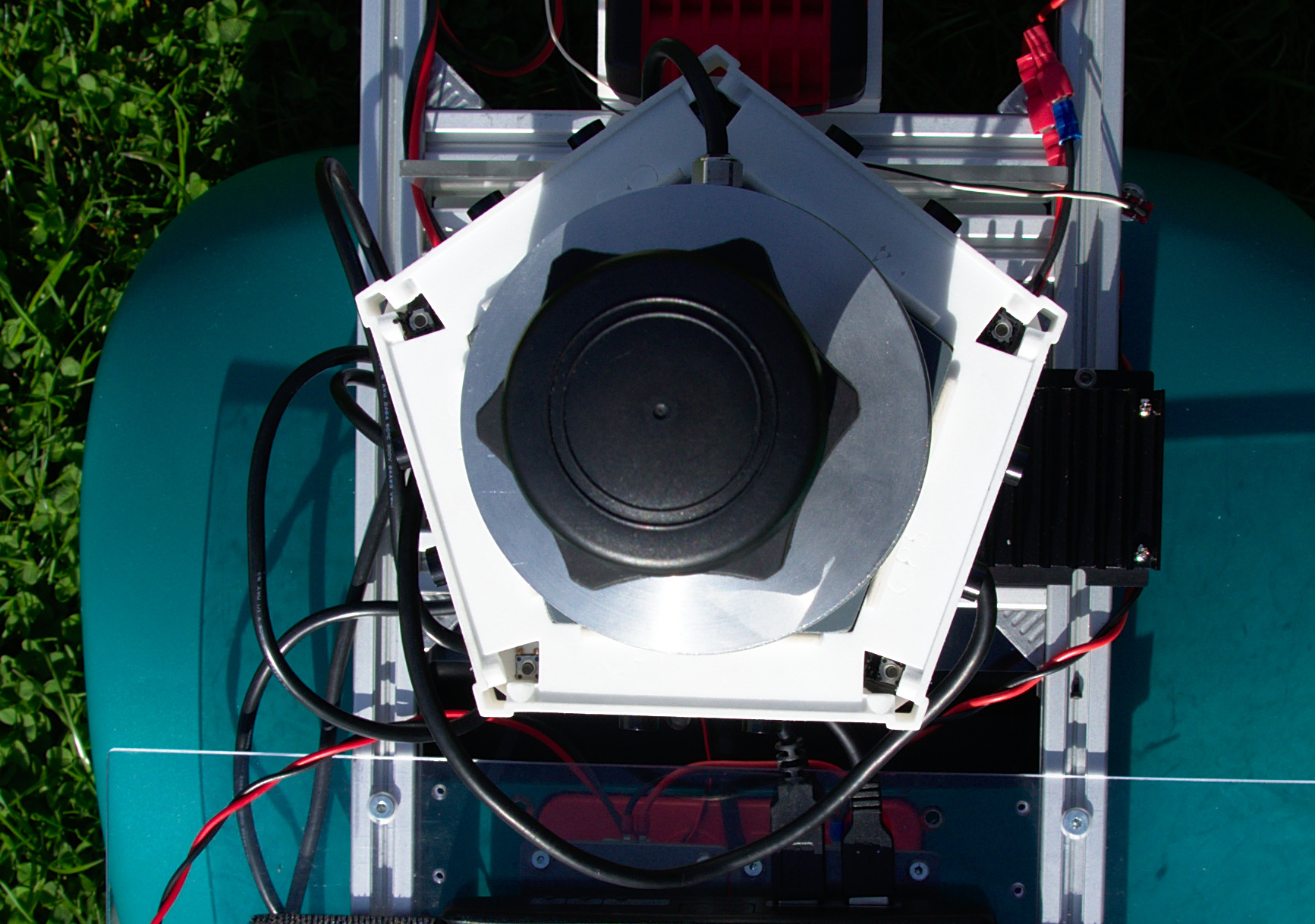}}
\caption{(a) Robot platform in the Wageningen garden of TrimBot2020 project. (b) Camera rig employed for the recording sessions.}
\label{fig:garden}
\end{figure}

We propose an extended version of the TB-Places data set~\cite{Leyva2019}, which consists of images recorded in the test gardens of the TrimBot2020 project~\cite{strisciuglio2018trimbot2020}. The gardens are located at the campus of the Wageningen University and Research (Netherlands) and at the Bosch Research Center in Renningen (Germany). 
The original TB-Places data set is composed of three sub sets, corresponding to two recording sessions that took place in Wageningen in 2016 and 2017 (W16 and W17), and one recording session in the garden in Renningen in 2017 (R17). The data was recorded using a modified Bosch Indego lawn mower robot, with a camera rig with 5 pairs of stereo cameras, which have $360\degree$ field of view (see Fig.~\ref{fig:garden}). The cameras in the rig are synchronized by means of an FPGA and record pictures at a resolution of $752 \times 480$ pixels~\cite{strisciuglio2018trimbot2020}. 
Each image has an associated ground truth camera pose in the garden reference system, recorded with a TopCon laser tracker and an inertial measurement unit (IMU). Additionally, we constructed a place recognition ground truth, labelling all the possible pairs of images indicating whether they depict similar or dissimilar scenes. For details about the labelling process, we refer the reader to~\cite{Leyva2019}.

\begin{table}[!t]
\small
    \centering
    \renewcommand{\arraystretch}{1}
    \setlength{\tabcolsep}{3mm}
    \begin{tabular}{l|c|c|c|c}
    \hline
    \textbf{Garden}     & \textbf{Set} & \textbf{\# imgs} & \textbf{\# similar pairs} & \textbf{\% similar pairs} \\ \hline \hline
    Wageningen & W16 & 40752 & 5.12M & 0.6168 \\ \hline
    Wageningen & W17 & 10948 & 330K  & 0.5441 \\ \hline
    Wageningen & W18 & 23043 & 1.03M  & 0.3877 \\ \hline
    Renningen  & R17 & 7999 & 150K & 0.4822 \\ \hline \hline
\end{tabular}
\vspace{3mm}
    \caption{Details on the composition of the extended TB-Places data set with the new W18 set of images. We report, for each subset, the number of image pairs labelled as similar and their percentage among all the possible image pairs.}
    \label{tab:dataset}
\end{table}

\begin{figure}[!t]
    \centering
    \subfloat[]{
\includegraphics[width=.75\textwidth]{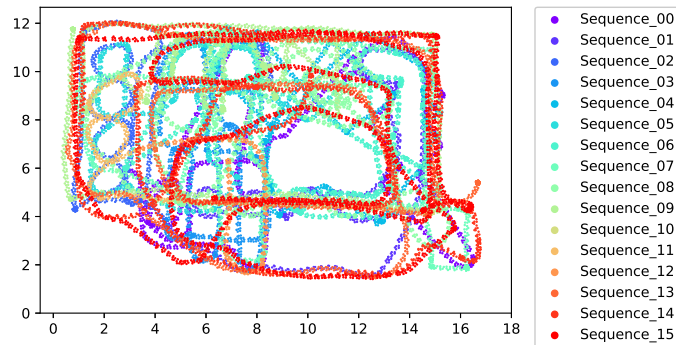}}
\\
    \subfloat[]{
\includegraphics[width=0.75\textwidth]{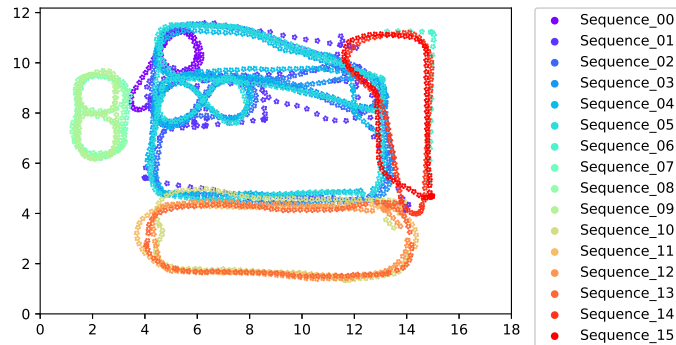}}
    \caption{Top-view of the trajectories followed by the robot during the recording sessions in the Wageningen garden in (a) 2018 (W18) and (b) 2017 (W17).}
    \label{fig:trajectories}
\end{figure}

\begin{figure}[!t]
\centering
        
\includegraphics[width=0.99\textwidth]{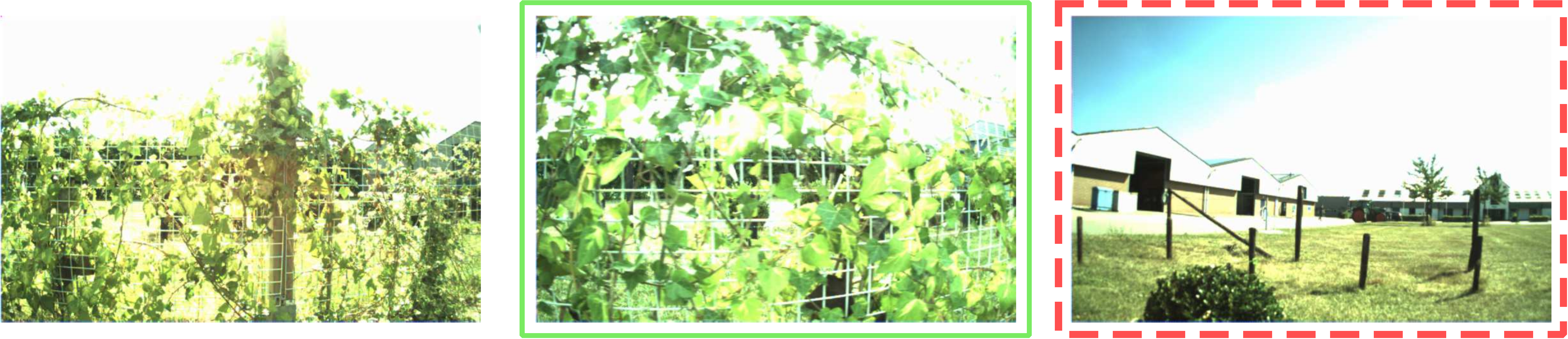}\\
\includegraphics[width=0.99\textwidth]{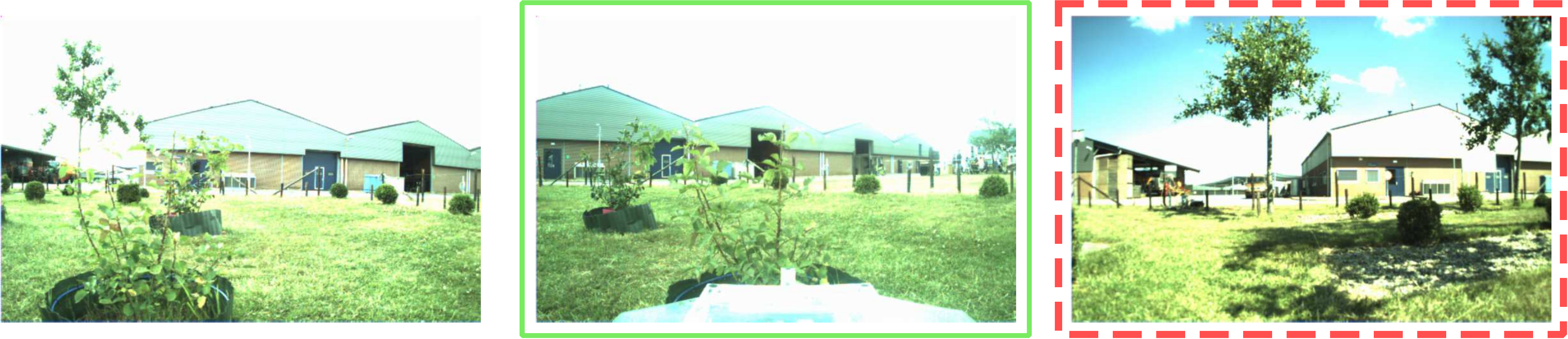}
    \\
\includegraphics[width=0.99\textwidth]{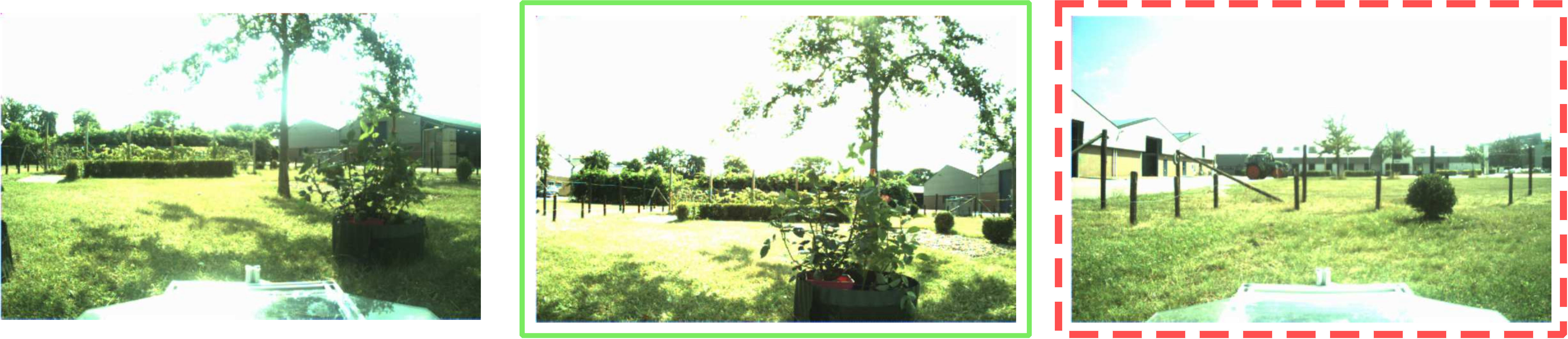}
    \\
\caption{Examples of TB-Places W18 data set. The left column shows reference images, while the center shows positive matches (green squares), and the right column shows negative matches (dashed red squares).}
\label{fig:tb_places_examples}
\end{figure}

We expanded the TB-Places data set with a new subset of 23K images (which we name W18), corresponding to a new recording session that took place in the garden in Wageningen during the summer of 2018. In Table~\ref{tab:dataset}, we report details on the composition of the data set. For the recording of the new set of images, the camera lenses where setup with an exposure value that allows to capture clearer details in the lower part of the image (i.e. closer objects and possible reference points) rather than in the upper part of the image (mostly sky). This provides visual localization algorithms with landmarks of finer quality for more precise camera pose estimation. The W18 set consists of 16 sequences, the trajectories of which are shown in Fig.~\ref{fig:trajectories}a. As displayed in Fig.~\ref{fig:trajectories}a and  Fig.~\ref{fig:trajectories}b, W18 data set covers parts of the environment that are not recorded in W17. Similarly to the rest of the data set, these additional images are provided with an associated camera pose in the garden reference system and a pair-wise similarity ground truth, which we constructed with the method proposed in~\cite{Leyva2019}. Some examples of images in the W18 set are included in Fig.~\ref{fig:tb_places_examples}.

\section{Evaluation}
\label{sec:experiments}
For the experiments, we used the W17 set as training set and the W18 set as test set. This setting allows to evaluate the robustness of image descriptors for place recognition against seasonal changes of the garden, color variations, as well as generalization to parts of the garden that are not seen during training.

We carried out experiments to evaluate the performance of state-of-the-art holistic image descriptors that we took as the baseline, and show that garden-specific descriptors can be developed by learning suitable representations using the garden images contained in the TB-Places data set. In the rest of the section, we provide details about the employed baseline descriptors and the learning process that we implemented.
\subsection{Performance measures}
We evaluated the performance of the considered descriptors (see Section~\ref{sec:baseline} for details) by computing the precision (P), recall (R) and $F_1$-score for the classification of pairs of images as depicting similar or dissimilar places:
\begin{equation*}
P=\frac{TP}{TP+FP}, R=\frac{TP}{TP+FN}, F_1=\frac{2\times P \times R}{R+P}
\end{equation*} where TP stands for true positives, TN is true negatives and FN means false negatives. A positive sample corresponds to a pair of images depicting the same scene. We present the results that we achieved in the form of a precision-recall curve, that we constructed by varying a threshold $t$ on the Euclidean distance of the computed image descriptors. All pairs with distance lower or equal to $t$ are classified as similar (positive).

Moreover, as an overall performance measure, we compute the Average Precision (AP), defined as: 
\begin{equation*}
AP =\sum _t(\text{R}_t-\text{R}_{t-1})P_n
\end{equation*} 

\subsection{Baseline}
\label{sec:baseline}

We selected three state-of-the-art holistic image descriptors as the baseline descriptors for evaluation of place recognition performance on the extended TB-Places data set. We considered the representation computed at the last layer of the ResNet-152~\cite{he2016deep}, DenseNet-161~\cite{huang2017densely} and  NetVLAD~\cite{Arandjelovic2017} architectures. For NetVLAD, we also evaluated the feature vectors computed at two intermediate layers, namely relu5\_2 and pool5. In the case of ResNet and DenseNet, the baseline models were trained on ImageNet [6], while the baseline NetVLAD model was trained on the Pittsburg30k data set~\cite{Torii2015pami}. When the output descriptor is three dimensional, we apply a \emph{AvgPool} operation, which outputs the average of the activation of each kernel, and a \emph{MaxPool}, which computes the maximum.
In Table~\ref{tab:descriptors} we report further details about the dimensionality of the considered image descriptors.

\begin{table}[!t]
\footnotesize
\centering
\setlength{\tabcolsep}{3mm}
\begin{tabular}{l|c|c|c}
\hline
\textbf{Model}                     & \textbf{Layer}                     & \textbf{Feature Pooling} & \textbf{Descriptor size}    \\ \hline\hline
\multirow{10}{*}{NetVLAD} & VLAD                      & -               & 4096               \\ \cline{2-4} 
                          & \multirow{3}{*}{AvgPool}  & -               & (512x7x7)=25088    \\ \cline{3-4} 
                          &                           & MaxPool         & 512                \\ \cline{3-4} 
                          &                           & AvgPool         & 512                \\ \cline{2-4} 
                          & \multirow{3}{*}{relu5\_2} & -               & (512x7x10)=35840   \\ \cline{3-4} 
                          &                           & MaxPool         & 512                \\ \cline{3-4} 
                          &                           & AvgPool         & 512                \\ \cline{2-4} 
                          & \multirow{3}{*}{pool5}    & -               & (512x7x10)=35840   \\ \cline{3-4} 
                          &                           & MaxPool         & 512                \\ \cline{3-4} 
                          &                           & AvgPool         & 512                \\ \hline
ResNet                    & AvgPool                   & -               & 2048               \\ \hline
\multirow{3}{*}{DenseNet} & \multirow{3}{*}{norm5}    & -               & (2208x7x11)=170016 \\ \cline{3-4} 
                          &                           & MaxPool         & 2208               \\ \cline{3-4} 
                          &                           & AvgPool         & 2208               \\ \hline
\end{tabular} \vspace{3mm}
\caption{Details on the descriptors that we considered in the benchmark analysis. }
\label{tab:descriptors}
\end{table}

\subsection{Learning garden-specific representations}
\begin{figure}[t!]
    \centering
    \includegraphics[width=.9\textwidth]{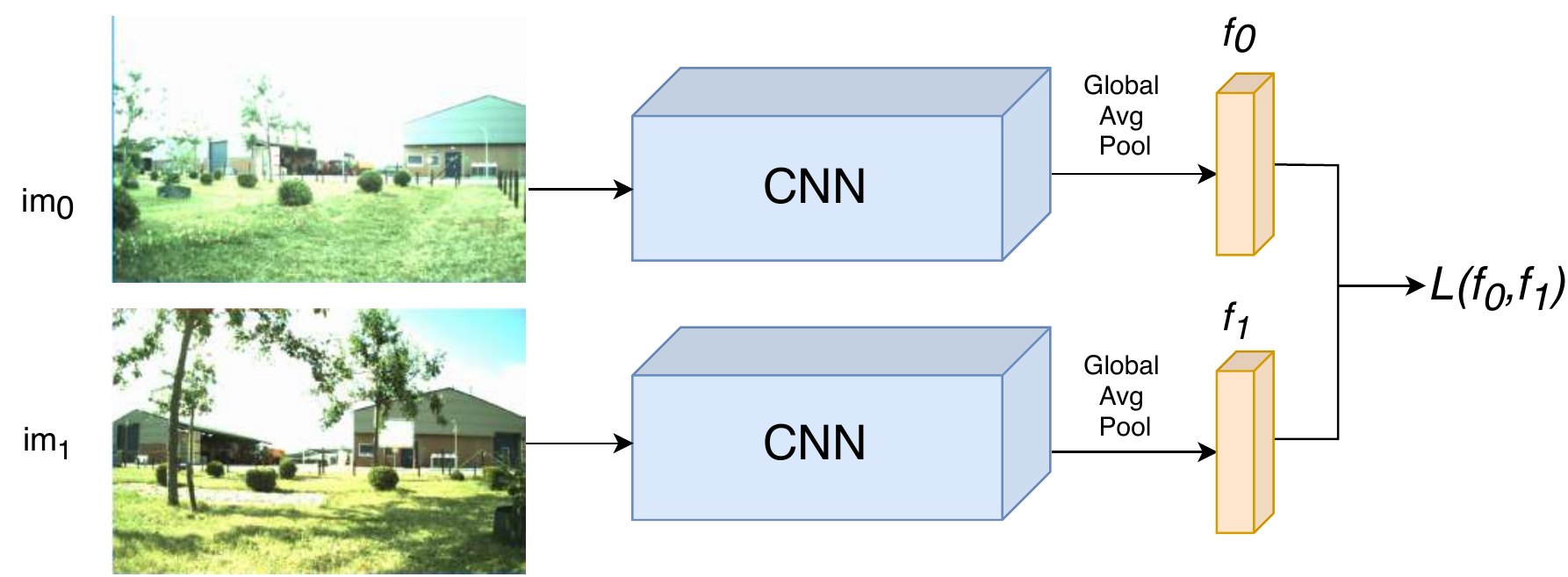}
    \caption{Sketch of the employed architecture for learning garden-specific descriptors for place recognition. We feed a pair of images to a siamese CNN architecture and compute their representation with a Global Average Pool layer. The training is guided by optimizing a contrastive loss function $L(f_0,f_1)$.}
    \label{fig:siamese}
\end{figure}

We employed a siamese CNN architecture where two backbone networks that share their weights to compute the descriptors of two input images (see Fig.~\ref{fig:siamese}). During training, the weights of the networks are updated so as to optimize a Contrastive Loss function $L(f_0,f_1)$ that compares the computed descriptors, $f_0$ and $f_1$. As initial conditions for the learning process, we selected the ResNet and DenseNet models pre-trained on ImageNet, which we consider good initialization for place recognition, and NetVLAD, which is pre-trained on the Pittsburgh30k data set. We learn garden-specific representations by training the considered models with images of the TB-Places data set.

For all models, we evaluate the descriptor that we compute by means of a \emph{global AvgPool} layer that we add in the end of the network. 
We use these descriptors to optimize a Contrastive Loss function $L$, formally defined as:
\[
 L(f_0, f_1) = \frac{1}{2N}\sum_{n=0}^{N-1} \left ( y^{(n)}d(f_0^{(n)}, f_1^{(n)})^2   + (1-y^{(n)}) \cdot\max(\alpha-d(f_0^{(n)}, f_1^{(n)}),0)^2\right)
\]
where  $d(f_{n0}, f_{n1}) = \left| \left| f_{n0} - f_{n1} \right| \right|_2 $ is the Euclidean distance between the feature vectors $f_0$ and $f_1$, $y_n$ is the ground truth label (0: dissimilar, 1: similar). The hyperparameter $\alpha$ is the loss margin, i.e. a threshold value of the descriptor distance above which a pair of images is not considered as depicting the same place. We set $\alpha$ as the value of the threshold that contributes to the highest $F_1$-score on the training set (W17). The selected values are reported in Table~\ref{tab:fts}. 

We trained the considered models, for 15 epochs, using the 10K images included in the W17 set, which  contain 330K image pairs with positive class label.
We set the batch size equal to 16 pairs and included in each training epoch the 330k positive pairs and an equal number of negative pairs, in order to avoid class unbalance while training.  The total amount of training iterations is thus approximately 620k.
We trained with Stochastic Gradient Descent, with one initial learning rate of 0.01, that was decreased by $10^{-1}$ every 5 epochs.

As stated in~\cite{Arandjelovic2017}, fine-tuning all the convolutional kernels in a CNN is not necessary to learn effective descriptors for visual place recognition. Early layers of a CNN, indeed, learn low level visual features (i.e. gabor-like filters)~\cite{Krizhevsky2012}, while upper deeper learn to detect more complex, task-specific features. We thus train the parameters of layers deeper layers, which compute semantically rich features. In particular, for the ResNet-152 and DenseNet-161 backbone CNNs, we train the last residual block (block5), and for the VGG network of NetVLAD, we trained the last convolutional block (conv5).
This choice allows to learn the value of a reduced set of parameters, i.e. only those of the deeper convolutional layers, simplifying the optimization problem and reducing the training time.

\begin{table}[!t]
\footnotesize
\centering
\setlength{\tabcolsep}{3mm}
\begin{tabular}{l|c|c}
\hline
\textbf{Model}              & \textbf{Feature length} & $\boldsymbol\alpha$ \\ \hline\hline
NetVLAD AvgPool + Global MaxPool & 512                     & 7.5             \\ \hline
ResNet152 AvgPool                & 2048                    & 12.7            \\ \hline
DenseNet norm5 + Global AvgPool  & 2208                    & 69.278          \\ \hline
\end{tabular} \vspace{3mm}
\caption{Details on the trained models. The second column displays the number of features of the learned holistic descriptors. The selected $\alpha$ values correspond to the threshold that produces the best $F_1$-score in the training set. }
\label{tab:fts}
\end{table}



\section{Results and discussion}
\label{sec:results}

\begin{table}[t!]
\centering
\footnotesize
\setlength{\tabcolsep}{2mm}
\begin{tabular}{l|c|c|c|c}
\hline
\multirow{2}{*}{\thead{\\\textbf{Descriptor}}}      & \multicolumn{2}{c|}{\textbf{Baseline}} & \multicolumn{2}{c}{\textbf{Trained}}                    \\ \cline{2-5} 
                                  & \textbf{W17}  & \textbf{W18}  & \textbf{\thead{W17 \\(train)}} & \textbf{\thead{W18 \\(test)}} \\ \hline\hline
NetVLAD VLAD                      & 0.1470        & \textbf{0.1383}        & -                    & -                   \\ \hline
NetVLAD AvgPool                   & 0.1438        & 0.1018        & 0.4627               & 0.1921              \\ \hline
NetVLAD AvgPool + Global AvgPool  & 0.1610        & 0.1264        & 0.7016               & 0.1230              \\ \hline
NetVLAD AvgPool + Global MaxPool  & 0.1215        & 0.0844        & 0.4273               & 0.0893              \\ \hline
NetVLAD relu5\_2                  & 0.1185        & 0.0812        & 0.4256               & 0.1470              \\ \hline
NetVLAD relu5\_2 + Global AvgPool &\textbf{ 0.1671}        & 0.1330        & 0.6978               & 0.1252              \\ \hline
NetVLAD relu5\_2 + Global MaxPool & 0.1241        & 0.0894        & 0.4052               & 0.0879              \\ \hline
NetVLAD pool5                     & 0.1169        & 0.0899        & 0.4087               & 0.0996              \\ \hline
NetVLAD pool5 + Global AvgPool    & 0.1590        & 0.1301        & 0.6749               & 0.1655              \\ \hline
NetVLAD pool5 + Global MaxPool    & 0.1323        & 0.0972        & 0.4230               & 0.1011              \\ \hline
ResNet152 AvgPool                 & 0.0991        & 0.0891        & 0.6616               & 0.1605              \\ \hline
DenseNet norm5 + Global AvgPool   & 0.1288        &  0.1078        & \textbf{0.7055}               & \textbf{0.2339 }             \\ \hline
DenseNet norm5 + Global MaxPool   &  0.1040       & 0.0724       & 0.6591               & 0.2318              \\ \hline
\end{tabular}
\vspace{3mm}
\caption{Achieved performance (Average Precision) in W17 and W18 data sets before (Baseline) and after training the models.}
\label{tab:results}
\end{table}

\begin{figure}[t!]
    \centering
        \subfloat[]{\includegraphics[width=.9\textwidth]{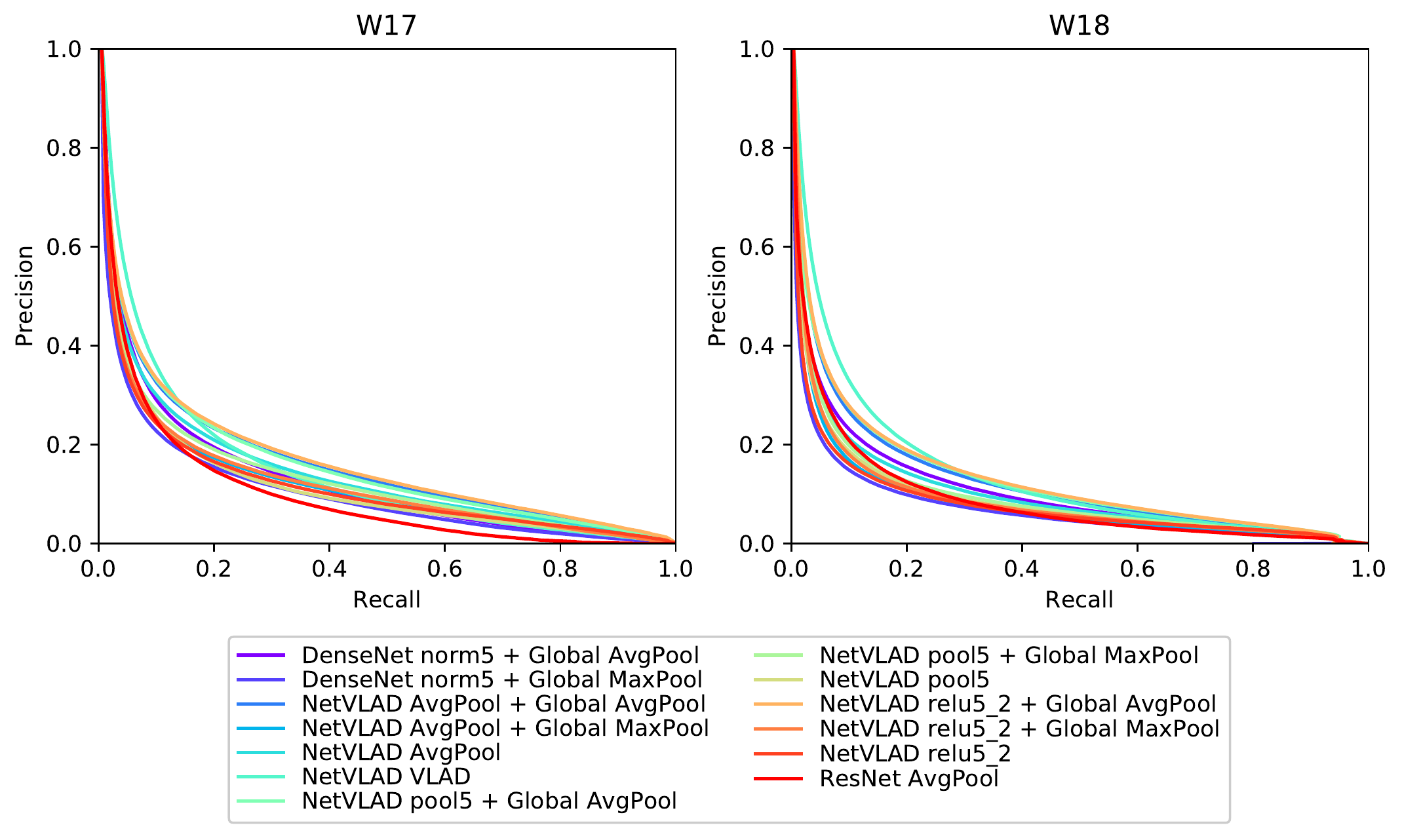}
    }
    \\
        \subfloat[]{
    \includegraphics[width=.9\textwidth]{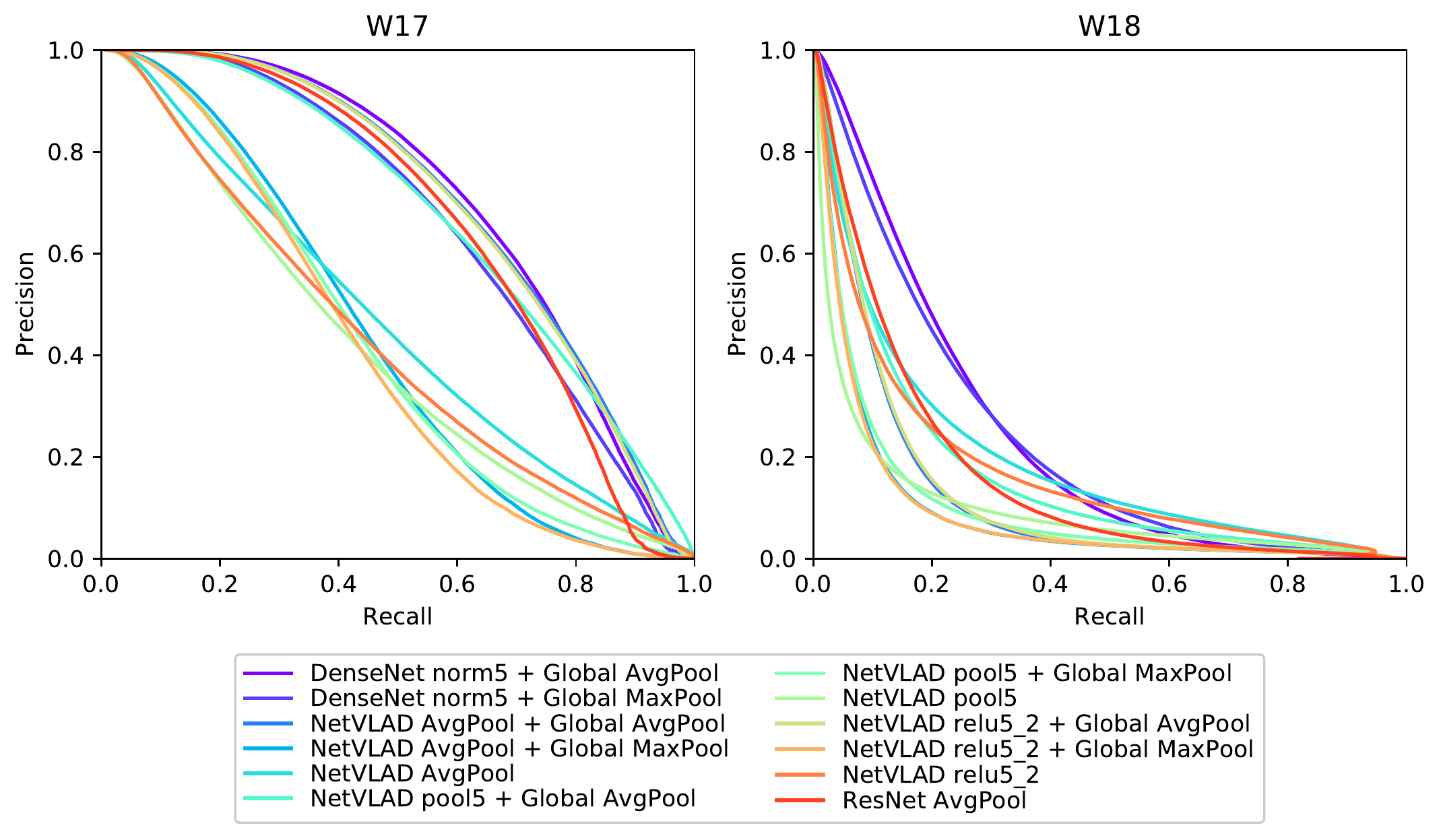}
    }\caption{Precision-recall curves achieved on train (W17) and test (W18) by the considered descriptors (a) before  and (b) after training the models.}
    \label{fig:pr_ft}
\end{figure}

\begin{figure}[t!]
    \centering
    \includegraphics[width=.6\textwidth]{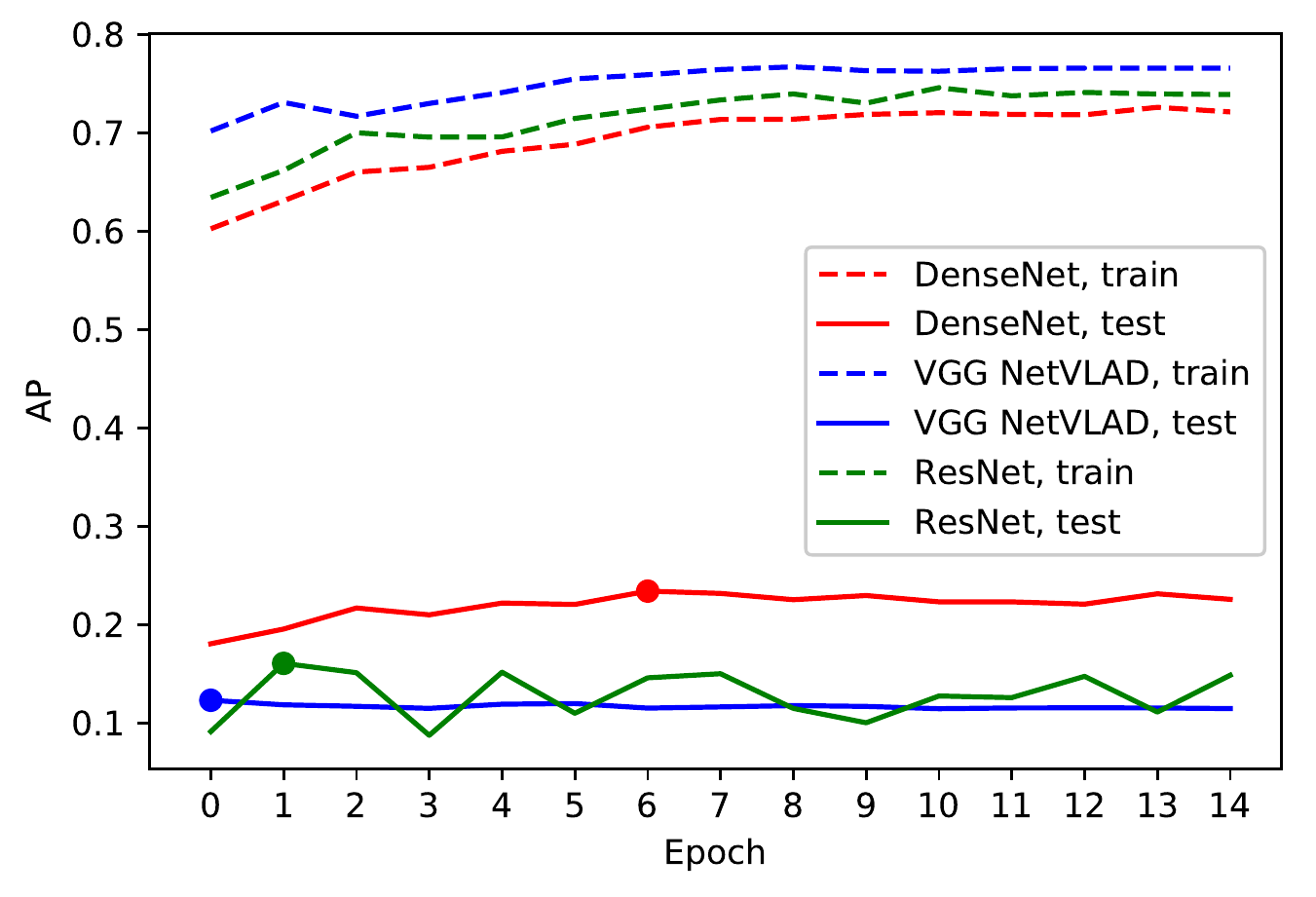}
    \caption{Train and test performance vs training epoch. The bullet indicates the best test AP achieved by each method, corresponding to epoch 6 for DenseNet, epoch 0 for NetVLAD VGG, and epoch 1 for ResNet.}
    \label{fig:epochs}
\end{figure}

In Table~\ref{tab:results}, we report the results that we achieved using the baseline descriptors and the ones that we learned from the images of the TB-Places data set. The highest value of the AP is obtained by NetVLAD both on the training set (W17) and on the test set (W18). More specifically, the descriptor computed with global average pooling on the relu5\_2 layer of NetVLAD achieved the best performance results (AP=0.1671) on the W17 set. On the W18 set, instead, the highest AP value (AP=0.1383) was obtained by using as image descriptor the output of the VLAD layer of NetVLAD.
After training the models using the siamese architecture described in Section~\ref{sec:experiments}, we observed an overall improvement of results as the networks are able to learn garden-specific features that contribute to improve the place recognition performance. The model that best performs after training is DenseNet, which achieves AP values of 0.7055 and 0.2339 on the train set (W17) and (W18) test set, respectively. In Fig.~\ref{fig:pr_ft}a and~\ref{fig:pr_ft}b, we show the Precision-Recall curves before and after training, respectively. All models improve their performance on the train and test set. The best results were obtained by DenseNet.

We observed that, although training the considered networks on the garden images contained in the TB-places data set contributes to improve place recognition results, the  resulting models suffer from a certain degree of specialization. In Fig.~\ref{fig:epochs}, we show a plot that illustrate the effect of the training process on the specialization of the models.
 In the case of VGG NetVLAD, this happens right after the end of the first epoch, while for ResNet it happens after the second epoch ($>$ 80K iterations). The DenseNet model reaches its best performance in epoch six. This indicates that the methods, although able to learn effective garden descriptors, have a tendency to overfit after a certain number of iterations. We conjecture that this is due to the fact that the W18 data set images are captured with modified exposure setting of the camera, and CNNs are very sensitive to variations in the input, producing unstable outputs~\cite{zheng2016improving,strisciuglio2019push}.



\section{Conclusions}
\label{sec:conclusions}
We proposed an extended version of the TB-Places data set, that includes more than 23k additional images recorded in a real garden of the Trimbot2020 project. The data set is designed to stimulate the benchmark of place recognition algorithms in garden environments.

We compared the performance of several CNN-based holistic image descriptors, namely those computed by ResNet, DenseNet and NetVLAD architecture, on the task of place recognition in garden scenes. We learned garden-specific image descriptors and demonstrated that garden-scene representations can be learned and improve the recognition results. However, their performance is affected by variations of the appearance of the environment due to changing light and weather conditions, or to color variations caused by modifications of exposure settings of the cameras. Thus, their generalization capabilities are limited. Therefore, the design of new robust image descriptors is required for effective visual localization or navigation systems for gardening and agricultural robotics.

\section*{Acknowledgements}
This work was funded by the European Horizon 2020 program, under the project TrimBot2020 (grant No. 688007).
\bibliographystyle{plain}

\end{document}